\documentclass[letterpaper, 10 pt, conference]{ieeeconf}
\IEEEoverridecommandlockouts
\usepackage{cite}
\usepackage{amsmath,amssymb,amsfonts}
\usepackage{algorithmic}
\usepackage{graphicx}
\usepackage{textcomp}
\usepackage{hyperref}
\usepackage{calc}  
\usepackage{multicol}
\usepackage{multirow}
\usepackage{enumitem} 
\usepackage[dvipsnames,table,xcdraw]{xcolor}
\usepackage[normalem]{ulem}
\useunder{\uline}{\ul}{}
\def\BibTeX{{\rm B\kern-.05em{\sc i\kern-.025em b}\kern-.08em
    T\kern-.1667em\lower.7ex\hbox{E}\kern-.125emX}}

\usepackage{subcaption}

\begin{document}

\title{Exploring the Cost of Interruptions in Human-Robot Teaming}



\author{Swathi Mannem$^{1}$, William Macke$^{1}$, Peter Stone$^{1,2}$, Reuth Mirsky$^{3}$
\thanks{$^{1}$Computer Science Department at The University of Texas at Austin
        {\tt\small \{swathimannem,wmacke, pstone\}@cs.utexas.edu}}%
\thanks{$^{2}$Sony AI, USA}
\thanks{        $^{3}$Computer Science Department at Bar Ilan University
        {\tt\small mirskyr@cs.biu.ac.il}
        }
%
}

\maketitle

\begin{abstract}
Productive and efficient human-robot teaming is a highly desirable ability in service robots, yet there is a fundamental trade-off that a robot needs to consider in such tasks. On the one hand, gaining information from communication with teammates can help individual planning. On the other hand, such communication comes at the cost of distracting teammates from efficiently completing their goals, which can also harm the overall team performance. In this study, we quantify the cost of interruptions in terms of degradation of human task performance, as a robot interrupts its teammate to gain information about their task. Interruptions are varied in timing, content, and proximity.
The results show that people find the interrupting robot significantly less helpful. However, the human teammate's performance in a secondary task deteriorates only slightly when interrupted. These results imply that while interruptions can objectively have a low cost, an uninformed implementation can cause these interruptions to be perceived as distracting. 
These research outcomes can be leveraged in numerous applications where collaborative robots must be aware of the costs and gains of interruptive communication, including logistics and service robots.
\end{abstract}


\section{Introduction}
The design of communicating agents and robots has been a fertile research area in multiagent systems and HRI~\cite{decker1987distributed,pagello1999cooperative}. 
When cooperating with teammates, there can be a fundamental tension between communication to understand their current objectives and distracting them from efficiently completing their objectives \cite{trafton2003preparing, monk2008effect, infantino2008human}.
Consider the case of a physician in a hospital and a service robot that can fetch specific medicine to patients according to the physician's current destination. If the robot is unsure which medicine to fetch next, it can interrupt the task execution and query the physician. Too many interruptions can distract the physician, increasing the physician's cognitive load and potentially even hindering the physician's performance, while too few can hinder the robot's performance as it is missing crucial information to be able to execute its task efficiently. Generally, while querying teammates about their plans can help one’s own planning, no real communication is ``cheap talk'', without any cost to the communicating parties \cite{crawford1998survey} thus it might harm the performance of one teammate or the team.

While the benefit of communication can be computed from an information theoretic perspective \cite{doshi2005particle, 
macke2021expected}, the extent to which communication is distracting to human teammates can only be estimated through human subject studies.
In this work, we specifically aim to understand human costs when communicating with a robot teammate. We explore how people perceive the cost of communication with a service robot and how this communication affects a human teammate's performance. We especially focus on a robot that can query its human teammate to elicit the human's goal. 
This paper further presents an experimental design for evaluating the cost of such interruptions using a collaborative task adapted from simulation to a physical setup. In this task, a human and a robot are a team, where the human is the leading party, and the robot needs to assist it. The human teammate needs to achieve a goal while performing a secondary memorization task. The robot can query the human, which causes an interruption, to understand the team's objectives better. We ran this study with varying timing, content, and proximity values of the interruptions to evaluate both the human participants' objective performance and their subjective experience regarding their performance and the collaboration with the robot.

This study shows that even when an interruption does not significantly deteriorate the human's performance in the secondary task, it is still perceived as interruptive by the human participants, causing them to rate the robot as more distracting and less helpful than a robot that does not query its teammate. The results of this study can directly lead to a better design of a robot that can judicially reason about the interruptive cost of a query, both in terms of objective cost and perceived cost, in human-robot teaming tasks. Additionally, this paper provides an experimental protocol that can be replicated and extended to suit studies on various other teaming tasks and human-robot interactions.

\section{Related Work}
\noindent \textbf{Cost of Interruptions}
The notion of quantifying the cost of communication stems from information theory, and it has often been used in the context of multi-agent systems \cite{decker1987distributed,pagello1999cooperative} and in HRI \cite{unhelkar2020decision}.
Specifically, Horvitz et al. \cite{horvitz2005balancing} investigated how a human perceives an interruption by a virtual office assistant and offered a bounded deferral approach to mitigate the disturbance of these interruptions. Rosenthal et al. \cite{rosenthal2011using} contribute a method of decision-theoretic experience sampling interruptions to learn when to automatically turn off and on the phone volume to avoid embarrassing phone interruptions.
An opposite use of interruptions as a desired artifact can be viewed in the Interruptions Skills Training and Assessment Robot. This robot allows adults with autism spectrum disorder (ASD) to practice handling interruptions to improve their employability \cite{ramnauth2022social}. The use of robots in such therapeutic contexts is supported by McKenna et al. \cite{mckenna2020sorry} who found that adults with ASD experience marginally less task disruption from a robot compared to a human.

Most closely related to our work is the work by Chiang et al. \cite{chiang2014personalizing}, who investigated the personalization of interruptions using reinforcement learning in the context of two distinct tasks: the human was reading a book, and the robot needed to interrupt about an unread message. This setup differs from the one in this research, where the robot is a teammate, and its objectives are highly coupled with the human's. 
Other than Chiang et al. who use verbal communication to interact with (and interrupt) the human, the HRI community mainly focuses on mitigating interruptions using shared autonomy \cite{mcgill2015team, gopinath2016human, laghi2018shared} or implicit communication, which are often less cognitively demanding than a conversation \cite{trovato2012development, mavridis2015review, allers2016evaluation}. Other work focuses on the cost the robot will incur for interacting with humans for help \cite{rosenthal2012someone}.

\begin{figure}[th]
    \centering
    \includegraphics[width=0.3\textwidth]{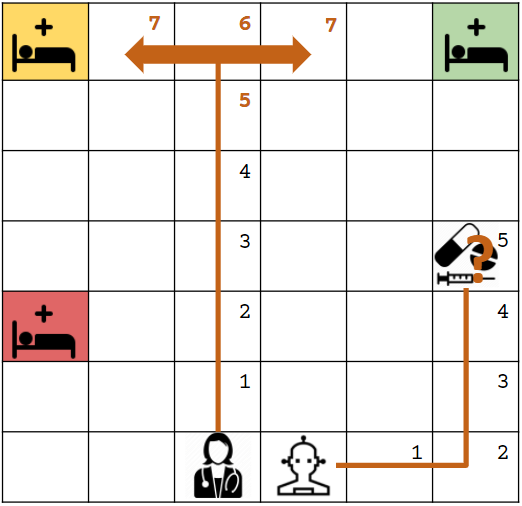}
    \caption{A depiction of the Tool Fetching domain. The robot needs to fetch the right medicine for the physician from the medical cabinet, according to the destination of the physician (image adapted from Suriadinata et al. \cite{suriadinata2021reasoning}).}
    \label{fig:tool-fetching}
\end{figure}

\noindent \textbf{Proxemics} In this work, we examine several potential variables that can affect the quality of the collaboration, specifically the interruptivness of a query: timing, content, and proximity. The latter is known to influence how people perceive an interaction and is investigated through the theory of proxemics \cite{hall1966hidden,mead2017autonomous}. According to Hall \cite{hall1966hidden},  there are four major proxemic zones, representing the horizontal distance between a person and other people around. 
Each zone holds a different expectation for interaction, thus querying at these varying distances might influence the perceived interaction.

\noindent \textbf{The Tool Fetching Domain} This work is inspired by a general multi-agent framework for Communication in Ad-hoc Teamwork (CAT) \cite{mirskypenny, macke2021expected}, in which the task is to design the behavior of one ad-hoc agent that needs to collaborate with previously unmet teammates. Consider the case depicted in Figure \ref{fig:tool-fetching} of a service robot that can fetch a specific medicine for a patient when the physician goes to that patient's room. If the physician starts to walk north, it might be unclear to the robot whether it should fetch the medicine for the patient in the yellow room or for the patient in the green room. In such a case, the robot might already reach the medical cabinet and still not know what medicine to fetch -- which can delay the total time of the team's task execution. While the physician could have chosen a route that made their destination clear from observation, previous research has shown that people do not reason about such legible action choices when working in a team, a result that justifies the use of interventions \cite{suriadinata2021reasoning}.
One such potential intervention is allowing the robot to query the physician, but such communication can also have a cost as it might interrupt the physician. To the best of our knowledge, this paper is the first that aims to quantify that cost and take it into account when deciding when to communicate.

\section{Exploring the Cost of Interruptions}
\label{sec:CostInt}
In the following work, we take the above tool-fetching use case and adapt it to the physical world. We implement a similar setup to Mirsky et al. \cite{mirskypenny}, where there are two agents in the environment: a worker (corresponding to the physician from the above use-case) and a fetcher (corresponding to the service robot). In the physical implementation of this setup, the fetcher's role is played by a physical robot, and a human participant plays the worker's role.
The human subject in this setup is given two simultaneous tasks. The main task is to navigate to one of the stations, described by a sequence of three shapes. The secondary task is memorizing a sequence of words, which the subject should write down on a note when reaching the station. Writing in each station requires a different pen, and the robot's role is to infer what station the human participant is heading toward and fetch the appropriate pen. To accomplish its goal, the robot can query the participant.
This setup is designed so the human participant leads the team effort, and the robot must reason about the participant's goal to fetch the right pen. If the robot cannot reach a conclusion regarding the participant's goal, it might not fetch the right pen on time. To avoid such mistakes or delays in the task, the robot might choose to interrupt the participant by querying about the right goal. The memorization task is used to evaluate and quantify how much the query will interrupt the person.

We explore several questions regarding the cost of queries from the human participant's perspective. We are interested both in objective effects as well as subjective effects on the human being queried:
\begin{description}
    \item [Q1] Will queries interrupt the memorizing task's quality?
    \item [Q2] Will queries interfere with the human's perception of the collaboration with the robot?
    \item [Q3] How will the distance between the agents at the time of the query affect the results? Specifically, will queries be perceived as more interrupting when asked in different zones as defined by proxemics principals \cite{hall1966hidden}? 
    \item [Q4] How will the timing of the question affect the results? Specifically, will queries be perceived as more interrupting in the beginning or middle of the task execution?
    \item [Q5] How will the content of the question affect the results? Specifically, will queries be perceived as more interrupting if the information gained from them is small?
\end{description}

\begin{figure*}
    \centering
     \begin{subfigure}[b]{0.40\textwidth}
     \centering
    \includegraphics[width=\textwidth]{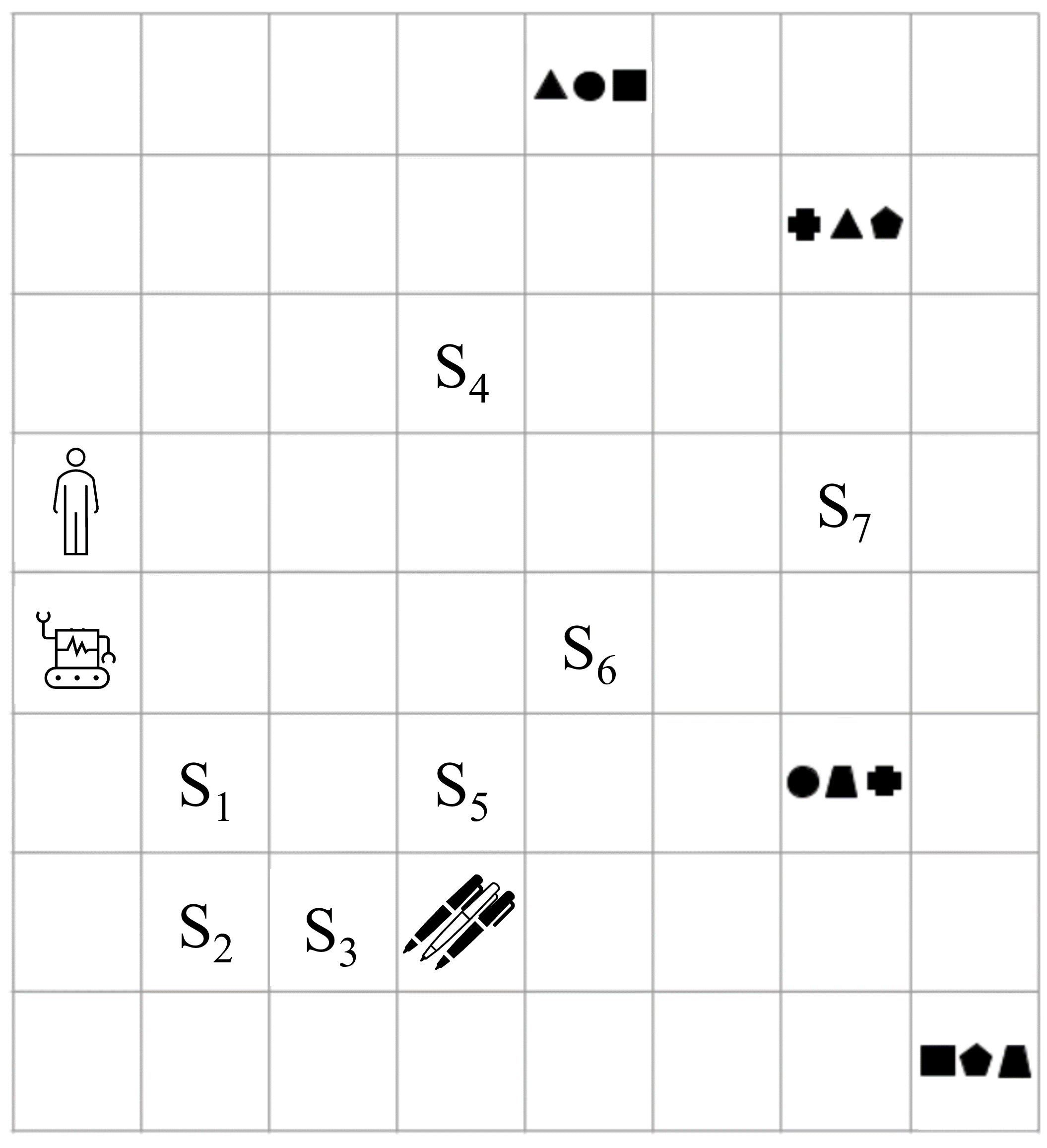}
    \end{subfigure}
    \hfill
     \begin{subfigure}[b]{0.58\textwidth}
         \centering
         \includegraphics[width=\textwidth]{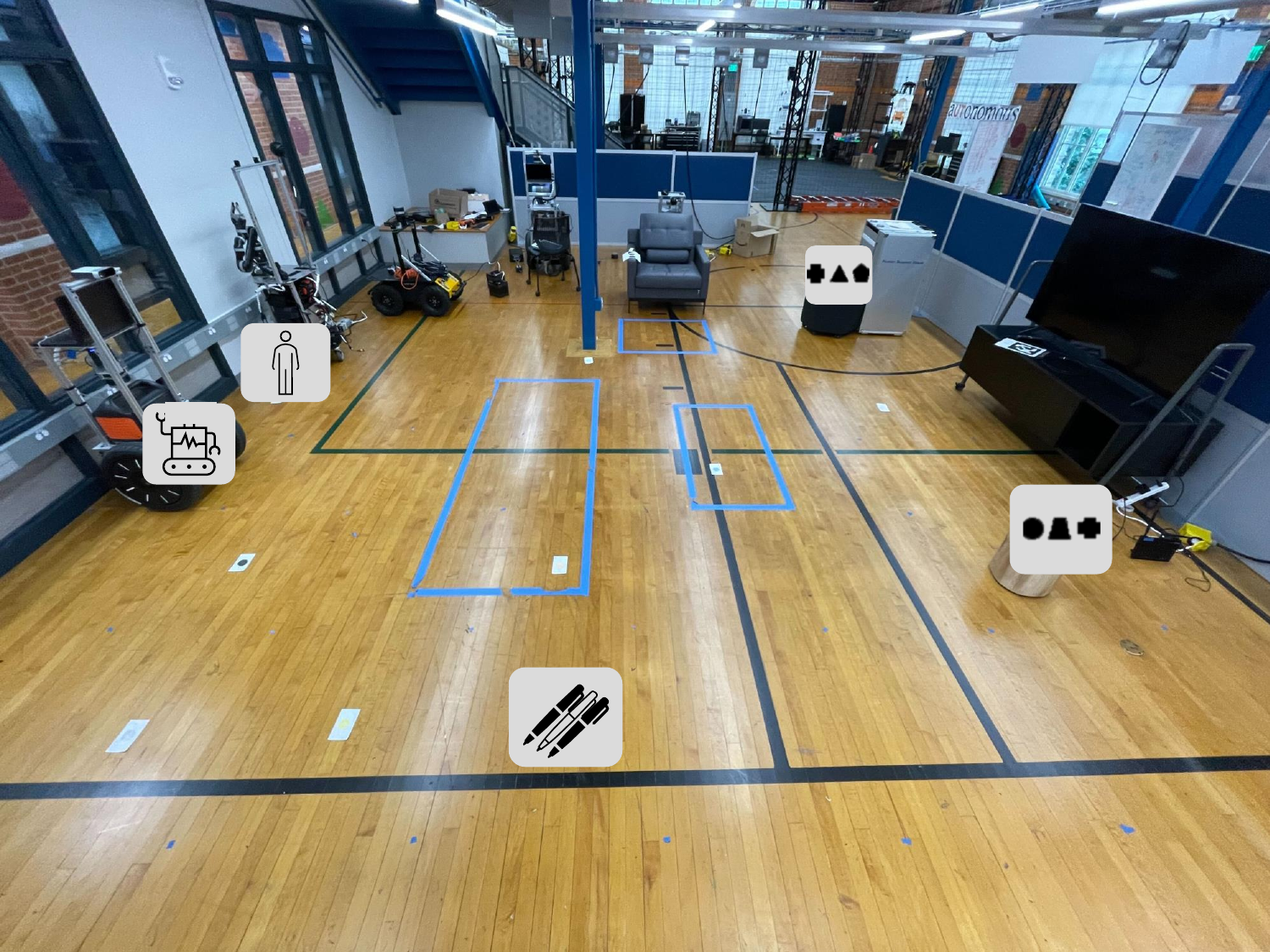}
    \end{subfigure}
        \caption{A depiction of the environment setup used in the experiment: an illustration (left) and the physical setup (right). The human and the robot icons represent the starting locations of the worker and fetcher, respectively; the pens represent the pens' location; the shapes represent the location of different workstations, and $S_i$ represent the intermediate locations where the teammates might be when a question is asked. The person, robot, pens, and two workstations are labeled on the physical setup to show the mapping between the configurations.}
        \label{fig:physical}
\end{figure*}

\section{Experimental Setup}
In the illustration of the setup in Figure \ref{fig:physical} (left), the human icon represents the initial location of the human participant, and the robot icon represents the location of the fetcher robot. We split the environment into a grid so each grid box is $1 \times 1$ meters (Figure \ref{fig:physical}). In the environment, there are four different stations, so a sequence of three shapes represents each station. Both agents move in the environment from one grid cell to a neighboring cell, and diagonal movement is not allowed. A confederate uses a metronome to set the pace of the movement, and the agents can travel from one grid space to the next on every beep. The first beep signals the start of the run, and the participant can start moving on the second beep. The pace of the metronome is set to 12 beats per minute, which provides enough time for both agents to move, reach a complete stop, and communicate a query and a response quickly if needed.

 The task of the worker is to reach one of the stations. As a secondary task, the worker is requested to memorize a sequence of related items (e.g., movie titles, cities, names, etc.), and write those down in a specific station. Following the design suggested in Mavrogiannis et al. \cite{mavrogiannis2019effects}, ``working'' in each station requires a pen of a different color to write the numbers. All pens are located in a single location (the black rectangle in the illustration), and the fetcher needs to choose the right pen and take it to the worker. 

We follow general proxemics principles and use the common values for interaction distances in the US \cite{hall1966hidden,mead2017autonomous}. We set an environment of $8 \times 8$ meters and we evaluate queries asked at four different distances between the agents (with the perspective proxemics zone stated in parentheses): 1.2m (far phase of personal space), 3m (social space), 4.2m (close phase of public space), and 5m (public space). 

Before the arrival of the participant, the environment is set up as shown in Fig. 2. The mobile robot used in this experiment is a Segway-based robot (the closest robot next to the left wall of the room) with a platform height of 110 cm such that a pen can be placed at a convenient height for a person to pick up from it. At each trial, the robot starts on the far left of the setup. There are two ways in which the robot can move: linearly forward and backward or in an angular direction by rotating left or right. 
The robot functions as a confederate in this experiment using a Wizard of Oz methodology (Woz), where it is being remotely controlled without the participant's knowledge. In addition, the participant believes that the robot reacts to the participant throughout the trial. However, the robot is teleoperated out of sight of the participant and has a predefined sequence of actions that it carries out according to a predefined protocol given to its operator. The robot's questions are vocalized using the Google translate text-to-speech program playing at $30\%$ volume from a speaker on the robot throughout all the trials. The environment's background noise is variable due to the space in which the experiment was conducted. However, it remains within an expected range in all trials; at times, talking occurred in the background, but there were no sudden or loud sounds.

\section{Experimental Procedure}
To avoid hypothesizing after the results are known (HARKing) \cite{hoffman2020primer}, this study was preregistered on OSF as \url{https://osf.io/gydp2}. 

 When the participant enters, they are first asked to read and fill out a consent form. Next, they read a statement of the purpose of the experiment, followed by a description of the scenario they are going through. 
 The participants are then directed to the experimental setup, where they are familiarized with the boundaries of the environment and the location of the stations. The participants are taken through a control round without the robot querying the participant. They are given an end station to go to, an intermediate location on the grid through which they must pass to control the distance between the worker and the fetcher when the question is asked, and a sequence of ten words within a category to memorize, with the category changing each trial. 
 The participant is then given 30 seconds to memorize this information and then directed to start walking towards the station at a pace set by a metronome, and upon arrival at the station directed to write down the words they remember on a sheet of paper. After each trial, the participant is requested to answer a questionnaire (Section \ref{sec:questionnaire}), and gets an additional minute to rest. The participant goes through this trial 15 more times with the addition of the robot into the experiment. 
 At the beginning of each trial, the robot starts moving at the same time as the participant and, upon arrival at a set intermediate location, asks the participant a predetermined question at a preset volume. As the participant answers, the robot continues to pick up the appropriate pen and then moves on to the working station. The robot's behavior is scripted to be optimal and to take the correct pen, whether the person replied correctly to the query or not. 
 Upon arrival at the station, the participants take the pen from the robot and write down the words they remember on a sheet of paper. 

\subsection{Memorization Task}
At each trial, the participant is asked to memorize ten words as a secondary task. These words are of the same category, such as movie titles or cities, but the category changes between trials to prevent confusion between the trials.  The categories assigned for each trial were randomized between participants. The content of the sequences was chosen according to previous research on working memory span tasks \cite{conway2005working} and tuned to a sequence length of ten using trial-and-error, so that it will be challenging yet feasible for the participant to recall. 

\subsection{Query Content}
\label{sec:questions_content}
As a reminder, the stations the participant can reach are represented using three shapes (e.g. cross-triangle-pentagon). There were three types of questions that were given to the participants across the different trials: 

\begin{enumerate}
    \item Are you going to the $\langle$\textit{three shape sequence}$\rangle$ station?
    \item Are you going to a station with a $\langle$\textit{single shape}$\rangle$?
    \item What station are you going to?
\end{enumerate}


These questions are used to evaluate if the content of the question has a significant effect on the worker's efficiency in performing the task and the accuracy of the response.
The first type of questions 
means that the fetcher asks about exactly one station in its full representation. It is a simple form of question but requires the worker to listen to a longer sequence of shapes. 
The second type of questions 
is slightly shorter in length, but answering such a question makes the worker think through the sequence of the goal station.
The last type of question consists of just one question: it is a more open-ended question, and while the response gives the information the fetcher requires for sure, it might have a higher cognitive load on the worker as it requires delineating the shape sequence of their target station while they still need to memorize the sequence of words in their secondary task. We contrast this question with the others to see if it is indeed more demanding on the worker.

In addition to the content of the different questions, we also investigate the impact of the question location and timing on the worker's efficiency in performing the task and the accuracy of the response. We evaluate four different querying distances between the agents: 1.2m, 3m, 4.2m, and 5m. For each distance, we also test the timing: whether the participant and robot are standing that far from one another in the beginning or middle of the task.

\subsection{Post-trial Questionnaire}
\label{sec:questionnaire}
The questionnaire distributed to participants at the end of each trial is an online form with 5-point Likert scale questions labeled from Strongly disagree to Strongly agree:
\begin{enumerate}
    \item Overall, my teammate was helpful.
    \item My teammate asking me the question was distracting.
    \item It was easy to convey the information to my teammate.
    \item The information I gave my teammate was effective in completing the task.
    \item I felt comfortable working with this team.
\end{enumerate}

This questionnaire is a modified PSSUQ \cite{lewis1992psychometric}, a validated post-study questionnaire for system usability, where instead of evaluating a system, we are asking the subject about the performance of the team. We report the Cronbach's alpha of this modified questionnaire ($\alpha=0.76$) to confirm that it maintains inner consistency.

\subsection{Test Configurations}
\label{sec:queries}
We mix between the content, timing, and distance to get 15 different configurations, as shown in Table \ref{tab:queries}. Each participant goes through all 15 configurations in a randomized order, as well as a first control trial with no queries asked by the robot. 


\begin{table}[ht]
\caption{The 15 configurations used in this experiment. \textbf{Worker} and \textbf{Fetcher} represent the location of the worker and the fetcher when a query is initiated (W/F in their original location or in one of the $S_i$ locations); \textbf{distance} and \textbf{timing} are the distance between the agent and the timing that the query is initiated; \textbf{query} is the question type asked from subsection \ref{sec:questions_content}; 
and \textbf{answer} is the expected answer of the worker.}
\footnotesize
\begin{tabular}{|l|l|l|l|l|l|l|}
\hline
\textbf{No.} & \textbf{Worker} & \textbf{Fetcher} & \textbf{Dist.} & \textbf{Timing} & \textbf{Query} & \textbf{Answer} \\ \hline
1            & W                                                                  & F                                                                   & 1.2               & Start       & 1                 & No                                                                 \\ \hline
2            & W                                                                  & F                                                                   & 1.2               & Start       & 2                 & Yes                                                                \\ \hline
3            & W                                                                  & F                                                                   & 1.2               & Start       & 3                & \includegraphics[width=0.6cm]{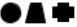}                                                                \\ \hline
4            & $S_1$                                                              & $S_2$                                                               & 1.2               & Middle          & 1                 & Yes                                                                \\ \hline
5            & $S_1$                                                              & $S_2$                                                               & 1.2               & Middle          & 2                 & Yes                                                                \\ \hline
6            & $S_1$                                                              & $S_2$                                                               & 1.2               & Middle          & 3                & \includegraphics[width=0.6cm]{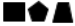}                                                                \\ \hline
7            & $S_4$                                                              & $S_5$                                                               & 3                 & Middle          & 1                 & No                                                                 \\ \hline
8            & $S_4$                                                              & $S_5$                                                               & 3                 & Middle          & 2                & Yes                                                                \\ \hline
9            & $S_4$                                                              & $S_5$                                                               & 3                 & Middle          & 3                & \includegraphics[width=0.6cm]{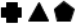}                                                                \\ \hline
10           & $S_6$                                                              & $S_3$                                                               & 4.2               & Middle          & 1                 & No                                                                 \\ \hline
11           & $S_6$                                                              & $S_3$                                                               & 4.2               & Middle          & 2                 & No                                                                 \\ \hline
12           & $S_6$                                                              & $S_3$                                                               & 4.2               & Middle          & 3                & \includegraphics[width=0.6cm]{Figures/spt.PNG}                                                                \\ \hline
13           & $S_7$                                                              & $S_3$                                                               & 5                 & Middle             & 1                 & No                                                                 \\ \hline
14           & $S_7$                                                              & $S_3$                                                               & 5                 & Middle             & 2                 & No                                                                 \\ \hline
15           & $S_7$                                                              & $S_3$                                                               & 5                 & Middle             & 3                & \includegraphics[width=0.6cm]{Figures/ctc.PNG}                                                                \\ \hline
\end{tabular}
\label{tab:queries}
\end{table}

\section{Results}
The experiment was conducted in accordance with the IRB of the University of Texas at Austin, IRB no. STUDY00002112. 
We recruited 30 participants, ranging in age from 18 to 22 years. As a within-subject design, all participants went through all 15 query setups in random order. Overall, each participant experienced every distance, timing, and content of the query condition in at least 3 trials. With this setup, the total number of trials is 450 trials with at least 90 instances of each evaluated condition. Using histogram plots, we verified that our data set is normally distributed to allow the use of ANOVA. The full distribution of trials can be seen in Table \ref{tab:queries}. 
The objective metrics considered for the analysis of the experiment are: precision, accuracy, and recall of the participant's memorization task, and the subjective metrics are helpfulness, distraction, and comfort.

\subsection{Objective Performance}
Recall that at each trial, the participant was asked to memorize a sequence of ten words from a specific category and write them down. As this process is not a full classification task, there are no items counted as ``True negatives (TN)'' -- all items are either ``True positives (TP), items that were correctly recalled; ``False positives (FP)'', items that were incorrectly added to the sequence; or ``False negatives (FN)'', meaning that the participant did not mention these items that appeared in the original sequence. Consequently, the following metrics were calculated:
\textit{Precision} is computed as the number of items the participant remembered correctly, divided by the number of items the participant wrote down (TP / TP + FP). \textit{Accuracy} is computed as the number of items the participant recalled correctly, divided by all correct items in the original sequence as well as the misspecified items (TP / TP + FP + FN). \textit{Recall} is computed as the number of items the participant remembered correctly, divided by the number of items in the original sequence (which is always ten). 

\begin{table*}[ht]
\caption{Mean values (and variance) of precision (P), accuracy (A), and recall (R) computed for each value of query distance, timing, content, and for the control with no query. For query content, ``Full'' are queries about a specific station (query type 1); ``Partial'' are queries about a single shape (query type 2); and ``What'' are the query ``What is your goal'' (query type 3). Significant values compared to the control with $p < 0.05$ have a green background.} 
\centering
\begin{tabular}{ll|llll|ll|lll|}
\cline{3-11}
                                                   &                                       & \multicolumn{4}{c|}{\textbf{Distance}}                                                                                                                                                                                     & \multicolumn{2}{c|}{\textbf{Timing}}                                               & \multicolumn{3}{c|}{\textbf{Content}}                                                                                                   \\ \cline{2-11} 
\multicolumn{1}{l|}{}                              & \multicolumn{1}{c|}{\textbf{Control}} & \multicolumn{1}{c|}{1.2}                           & \multicolumn{1}{c|}{3}                             & \multicolumn{1}{c|}{4.2}                           & \multicolumn{1}{c|}{5}                                      & \multicolumn{1}{c|}{Beginning}                     & \multicolumn{1}{c|}{Middle}   & \multicolumn{1}{c|}{Full}                          & \multicolumn{1}{c|}{Partial}                       & \multicolumn{1}{c|}{What}     \\ \hline
\multicolumn{1}{|l|}{}                             &                                       & \multicolumn{1}{l|}{}                              & \multicolumn{1}{l|}{}                              & \multicolumn{1}{l|}{}                              & \cellcolor[HTML]{DCEFD6}                                    & \multicolumn{1}{l|}{}                              &                               & \multicolumn{1}{l|}{}                              & \multicolumn{1}{l|}{}                              &                               \\
\multicolumn{1}{|l|}{\multirow{-2}{*}{\textbf{P}}} & \multirow{-2}{*}{.979 (.002)}         & \multicolumn{1}{l|}{\multirow{-2}{*}{.971 (.001)}} & \multicolumn{1}{l|}{\multirow{-2}{*}{.978 (.001)}} & \multicolumn{1}{l|}{\multirow{-2}{*}{.984 (.001)}} & \multirow{-2}{*}{\cellcolor[HTML]{DCEFD6}{\ul .957 (.003)}} & \multicolumn{1}{l|}{\multirow{-2}{*}{.973 (.001)}} & \multirow{-2}{*}{.972 (.001)} & \multicolumn{1}{l|}{\multirow{-2}{*}{.972 (.001)}} & \multicolumn{1}{l|}{\multirow{-2}{*}{.974 (.001)}} & \multirow{-2}{*}{.969 (.001)} \\ \hline
\multicolumn{1}{|l|}{}                             &                                       & \multicolumn{1}{l|}{}                              & \multicolumn{1}{l|}{}                              & \multicolumn{1}{l|}{}                              &                                                             & \multicolumn{1}{l|}{}                              &                               & \multicolumn{1}{l|}{}                              & \multicolumn{1}{l|}{}                              &                               \\
\multicolumn{1}{|l|}{\multirow{-2}{*}{\textbf{R}}} & \multirow{-2}{*}{.753 (.024)}         & \multicolumn{1}{l|}{\multirow{-2}{*}{.749 (.014)}} & \multicolumn{1}{l|}{\multirow{-2}{*}{.756 (.011)}} & \multicolumn{1}{l|}{\multirow{-2}{*}{.754 (.014)}} & \multirow{-2}{*}{.758 (.013)}                               & \multicolumn{1}{l|}{\multirow{-2}{*}{.758 (.117)}} & \multirow{-2}{*}{.752 (.104)} & \multicolumn{1}{l|}{\multirow{-2}{*}{.757 (.014)}} & \multicolumn{1}{l|}{\multirow{-2}{*}{.749 (.014)}} & \multirow{-2}{*}{.754 (.012)} \\ \hline
\multicolumn{1}{|l|}{}                             &                                       & \multicolumn{1}{l|}{}                              & \multicolumn{1}{l|}{}                              & \multicolumn{1}{l|}{}                              &                                                             & \multicolumn{1}{l|}{}                              &                               & \multicolumn{1}{l|}{}                              & \multicolumn{1}{l|}{}                              &                               \\
\multicolumn{1}{|l|}{\multirow{-2}{*}{\textbf{A}}} & \multirow{-2}{*}{.517 (.062)}         & \multicolumn{1}{l|}{\multirow{-2}{*}{.536 (.025)}} & \multicolumn{1}{l|}{\multirow{-2}{*}{.542 (.028)}} & \multicolumn{1}{l|}{\multirow{-2}{*}{.556 (.038)}} & \multirow{-2}{*}{.514 (.024)}                               & \multicolumn{1}{l|}{\multirow{-2}{*}{.549 (.034)}} & \multirow{-2}{*}{.534 (.020)} & \multicolumn{1}{l|}{\multirow{-2}{*}{.534 (.025)}} & \multicolumn{1}{l|}{\multirow{-2}{*}{.522 (.029)}} & \multirow{-2}{*}{.555 (.027)} \\ \hline
\end{tabular}
\label{tab:objective}
\end{table*}

Table \ref{tab:objective} shows the mean and variance values of each metric in each of the conditions evaluated. As seen in this Table, only in one condition (a 5m distance between the agents) was there a significant difference in precision from the control. Additionally,  a one-way ANOVA was conducted between the different memorization categories to test whether the difficulty of recalling the items of one category was significantly different than the other categories. The ANOVA results show a significant difference between the category means for precision ($F_{1,15} = 2.659$, $p < 0.001$), accuracy ($F_{1,15} = 3.887$, $p < 0.001$), and recall ($F_{1,15} = 5.331$, $p < 0.001$). Specifically, the Elements and Technology were the most challenging for the participants to memorize. The full list of the sequences for the memorization task can be viewed in the Appendix.

\begin{figure}[ht]
     \centering
     \begin{subfigure}[b]{0.40\textwidth}
         \centering
         \includegraphics[width=\textwidth]{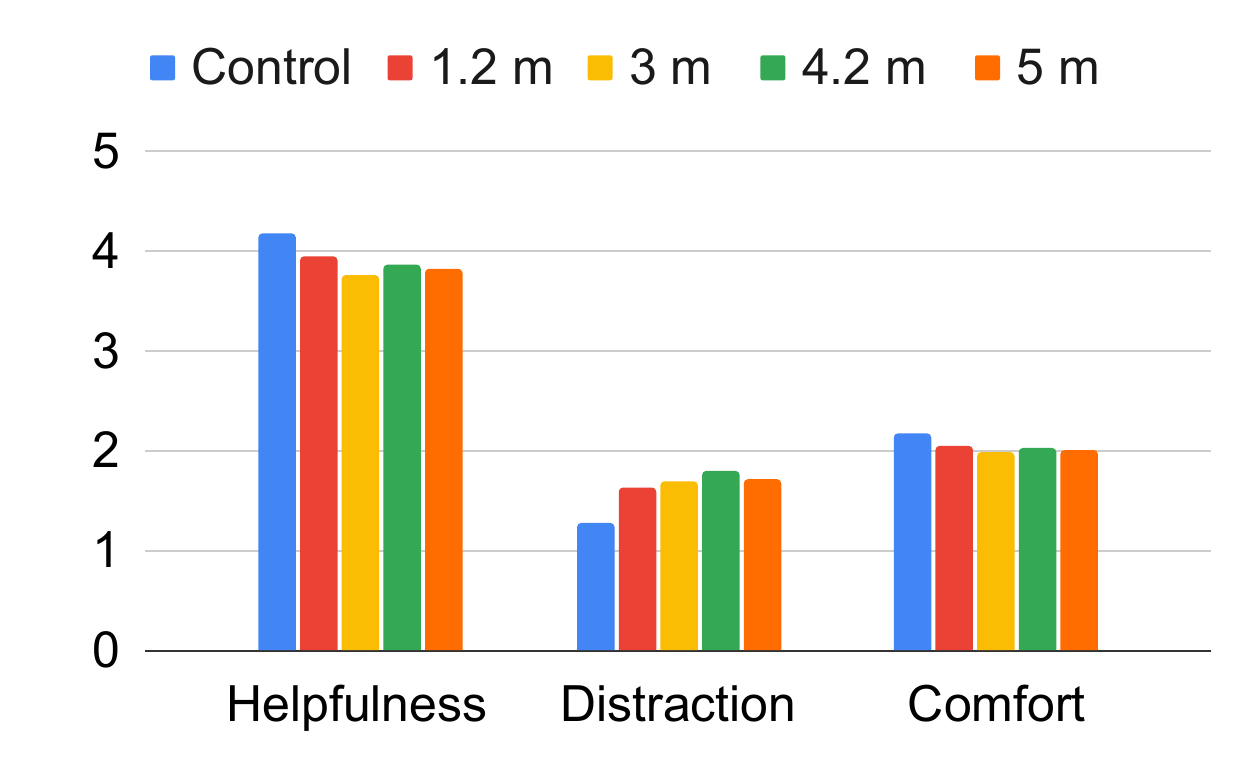}
         \caption{Proximity}
         \label{fig:proximity_survey}
     \end{subfigure}
     \hfill
     \begin{subfigure}[b]{0.40\textwidth}
         \centering
         \includegraphics[width=\textwidth]{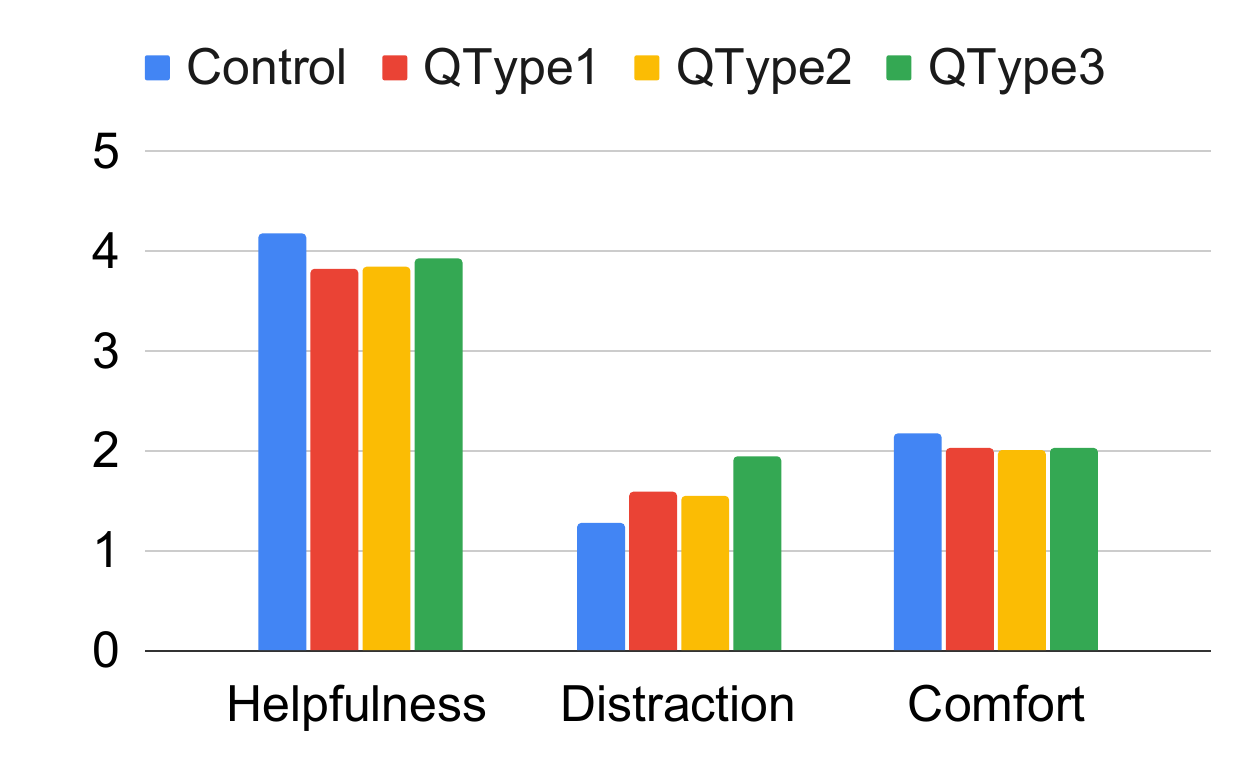}
         \caption{Query Type}
         \label{fig:QType_survey}
     \end{subfigure}
     \hfill
     \begin{subfigure}[b]{0.40\textwidth}
         \centering
         \includegraphics[width=\textwidth]{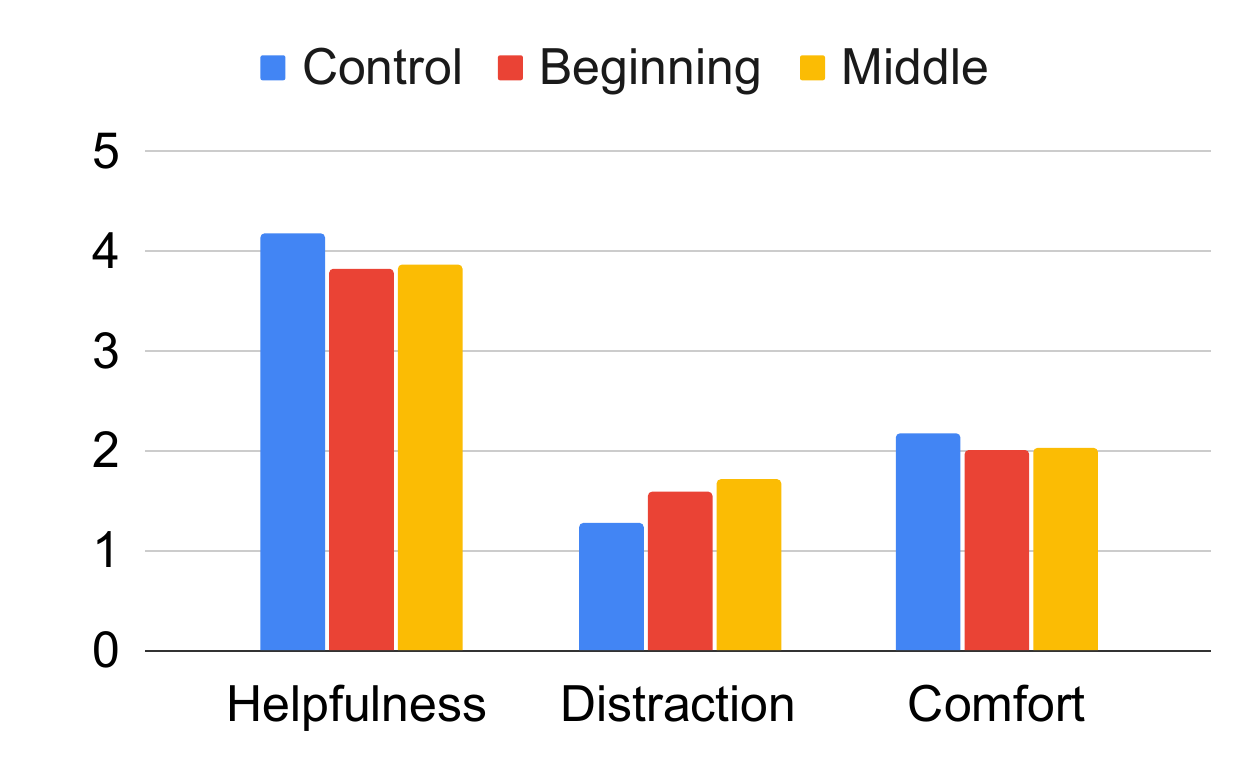}
         \caption{Timing}
         \label{fig:timing_survey}
     \end{subfigure}
    \caption{Questionnaire average responses according to proximity, query type, and timing.}
    \label{fig:questionnaire}
\end{figure}

\subsection{Questionnaire Results}
The participants filled out the questionnaire immediately after each trial. \textit{Helpfulness} is measured using the average of questions 1 and 4 from Section \ref{sec:questionnaire}. \textit{Distraction} is measured using the average of question 2 and the inverse of question 3. \textit{Comfort} is measured using question 5. We use ANOVA tests with Bonferroni corrections when reporting significance. In all conditions, the robot was considered helpful ($3.89 \pm 1.039$), slightly distracting ($1.67 \pm 0.98$), and the interaction was not very comfortable ($2.04 \pm 0.58$).



\noindent\textbf{Helpfulness:} For query distance, there was a significant difference between the control and all distances ($p < 0.01$). For timing, there was a significant difference between the control and all query timings ($p<0.01$). For content, there was a significant difference between the control and all query types ($p < 0.01$). 

\noindent\textbf{Distraction:} For distance, there was a significant difference between the control and 5m ($p < 0.01$).  For query timing, there was not a significant difference between the control and the query at the beginning ($p=0.23$) or the query in the middle ($p=0.03$). For query content, there was a significant difference between the control and query type 3 ($p<0.01$).

\noindent\textbf{Comfort:} There was no significant difference between the control and any of the distance, timing, or content values. 

\section{Discussion}


We now answer the questions posed earlier in Section~\ref{sec:CostInt}. First, regarding \textbf{Q1} our results do not support that querying had any significant detriment on the ability of the human collaborator to complete the specified task. ANOVA tests between the various querying experiments and the control did not show a significant difference between the precision, recall, or accuracy in the trials, with 1 exception out of 27 experimental conditions. 
A potential explanation for this lack of significance could be attributed to the difficulty of the memorization task -- if the dynamic range of performance in the task is not wide enough (e.g. if the task is always very challenging or very easy), then no difference would be observed between the different conditions. 

However, this lack of objective cost to the initiated queries does not mean that they incur no cost at all: regarding \textbf{Q2}, querying did have a significant impact on the participants' perception of how successful the collaboration was. While the objective performance of the secondary task remained relatively unaffected, the results show a significant difference in the participants' perception of how \textit{helpful} and \textit{distracting} the robot teammate was: many querying conditions were found to cause the participant to experience the robot as less helpful and more distracting, compared to the control trial without a query. We note that the effect size in these tests is often less than one point out of the 5-point scale. We conjecture that this small effect size stems from the short interactions, where the difference between the conditions is the query-and-answer part, which lasts about 5 seconds.

Interestingly, regarding \textbf{Q3-Q5}, the query timing did not seem to have any significant impact on either the objective results or the participants' perception of distraction. The content of the query only had a significant effect on the perceived distraction caused by the robot when the query was of type 3 (``What is your goal'') but not when it was of type 1 or 2 (``Is your goal ...''). This result supports the hypothesis in Section \ref{sec:questions_content} that this question type will be more demanding on the worker. Additionally, significance tests demonstrated that querying only had a significant detriment on participants' perception of distraction when the robot was querying from 5 meters away. A common assumption about interruptive queries is that they should be avoided when possible and only used when needed. However, this result offers a different prioritization: the robot should query when close by, even if it might not be necessary, to avoid the risk of querying later in the interaction when the teammate is further away.

We follow to report some limitations of this study. The participants are students at a university level in mostly STEM fields. This experience makes them inclined to be better at working with or memorizing numbers which could positively skew the results as compared to testing with a random sample of the population. The number of participants also limits the study's accuracy as it is not a perfectly representative sample of the population. The environment in which the study is being conducted varies slightly from trial to trial in the amount of background noise and the people moving around outside the experimental setup, both of which could be potential distractions to the participants. The robot's traits, such as vocal volume and morphology, might also have some unmeasured influence on the results of this study.   Lastly, to accustom the participants to the task, we always presented the control condition to them first.  Consequently, the participants may have been more alert in this first trial and thus achieved better results.  To mitigate this concern, we gave participants time to rest between the trials, which has worked since the objective measures showed no significant difference from the control. Nonetheless, it may be worth randomizing the sequencing of this control condition in any follow-up experimentation.


\section{Conclusion}
This paper investigates the effect of interruptive communication in human-robot collaboration. Specifically, it presents an experimental setup to evaluate the objective and perceived costs of interruptive queries initiated by the robot. The examined variables include proximity, timing, and content of a query initiated by a robot.
The results show that even when the human’s performance in a secondary task deteriorates only slightly when interrupted, the interrupting robot was perceived as significantly less helpful and more distracting than a non-interrupting robot. This outcome implies that the way humans perceive the cost of the interruption is not true to its objective cost, and both should be acknowledged and measured independently. This result also emphasizes the need to design the attributes of an interruptive query properly so that it provides the robot with the desired information and improves the overall team performance while not being perceived by human teammates as distracting.

One approach to directly reduce the perceived interruption cost was investigated in previous work on politeness \cite{srinivasan2016help}, yet other alternatives can be considered. One such potential attribute that arises from the results is an informed use of the robot's distance from the human to mitigate the perceived interruption. This attribute could be further tuned and optimized using proxemics principals \cite{mead2017autonomous}, e.g. by adjusting the pose and speech volume of the robot to suit the expectation of its human teammate better. 
Moreover, the experimental design proposed in this work can be used to investigate additional approaches to mitigate the perceived interruption cost in human-robot teaming tasks.
Another potential research avenue is leveraging shared attention to better react to a person who is distracted \cite{lombardi2022icub}. Alternatively, explainable AI (XAI) can be used to mitigate the perceived interruption \cite{Alshehri23}. 

Lastly, a complementary research problem is the quantification of the \textit{gain} from an initiated query -- some existing work that focuses on estimating the potential gain \cite{doshi2005particle, macke2021expected} could be combined with the outputs of this study to construct a collaborative robot that judicially reasons about the trade-off between the costs and gains of a query.

\section*{Acknowledgments}
This work has taken place in the Learning Agents Research Group (LARG) at the Artificial Intelligence Laboratory, The University of Texas at Austin.  LARG research is supported in part by the National Science Foundation (FAIN-2019844, NRT-2125858), the Office of
Naval Research (N00014-18-2243), Army Research Office (E2061621), Bosch, Lockheed Martin, and Good Systems, a research grand challenge at the University of Texas at Austin.  The views and conclusions contained in this document are those of the authors alone.  Peter Stone serves as the Executive Director of Sony AI America and receives financial compensation for this work.  The terms of this arrangement
have been reviewed and approved by the University of Texas at Austin in accordance with its policy on objectivity in research.
Research conducted in the Goal Optimization using Learning and Decision-making (GOLD) lab at Bar Ilan University is part of the HRI consortium supported by the Israel Innovation Authority.

\bibliography{main}
\bibliographystyle{abbrv}

\appendix

The participants were asked to memorize sequences of ten words in each trial. Below all of the categories with their respective sequences are listed:
\begin{description}[leftmargin=!,labelwidth=\widthof{\bfseries Measurements}]
    \item [Cities] Dallas, Seattle, Chicago, Austin, Orlando, Atlanta, Denver, Houston, Boston, Phoenix.
    \item [Fruits] Apple, Banana, Grape, Kiwi, Pineapple, Melon, Mango, Pear, Guava, Orange.
    \item [Elements] Zinc, Copper, Oxygen, Nitrogen, Sodium, Calcium, Helium, Neon, Nickel, Sulfur.
    \item [Vegetables] Carrot, Broccoli, Peas, Lettuce, Cauliflower, Celery, Eggplant, Pepper, Cabbage, Potato.
    \item [Countries] Mexico, Japan, France, Nigeria, India, Argentina, Ethiopia, Canada, Pakistan, China.
    \item [Technology] Apple, Samsung, Huawei, Toshiba, Sony, Intel, Microsoft, Dell, Cisco, Oracle.
    \item [School Subjects] Math, Gym, Business, Art, Biology, History, Chemistry, Music, Health, Government.
    \item [Body parts] Hands, Feet, Head, Eyes, Legs, Hair, Face, Arms, Toes, Spine.
    \item [Sports] Soccer, Football, Volleyball, Track, Baseball, Tennis, Swimming, Dance, Cricket, Rugby.
    \item [Furniture] Rug, Table, Desk, Counter, Shelf, Bed, Chair, Stool, Sink, Couch.
    \item [Animals] Dog, Bird, Snake, Cat, Wolf, Deer, Bear, Pig, Horse, Fish.
    \item [Clothing] Shirt, Belt, Scarf, Tie, Shoes, Vest, Scarf, Socks, Bracelet, Shorts.
    \item [Measurements] Inch, Gram, Meter, Second, Watt, Yard, Joule, Liter, Kelvin, Mole.
    \item [Flavors] Chocolate, Pistachio, Mint, Butterscotch, Vanilla, Coffee, Neapolitan, Caramel, Hazelnut, Sherbet.
    \item [Instruments] Piano, Flute, Trombone, Saxophone, Drums, Clarinet, Oboe, Guitar, Trumpet, Piccolo.
    \item [Plants] Grass, Mushroom, Tree, Fungus, Vine, Mold, Flower, Moss, Cactus, Weed.
\end{description}

\end{document}